# Phenotypical Ontology Driven Framework for Multi-Task Learning

Mohamed Ghalwash, Zijun Yao, Prithwish Chakraborty, James Codella, Daby Sow

*Abstract*—Despite the large number of patients in Electronic Health Records (EHRs), the subset of usable data for modeling outcomes of specific phenotypes are often imbalanced and of modest size. This can be attributed to the uneven coverage of medical concepts in EHRs. In this paper, we propose **OMTL**, an Ontology-driven Multi-Task Learning framework, that is designed to overcome such data limitations. The key contribution of our work is the effective use of knowledge from a predefined well-established medical relationship graph (ontology) to construct a novel deep learning network architecture that mirrors this ontology. It can effectively leverage knowledge from a well-established medical relationship graph (ontology) by constructing a deep learning network architecture that mirrors this graph. This enables common representations to be shared across related phenotypes, and was found to improve the learning performance. The proposed **OMTL** naturally allows for multi-task learning of different phenotypes on distinct predictive tasks. These phenotypes are tied together by their semantic distance according to the external medical ontology. Using the publicly available MIMIC-III database, we evaluate **OMTL** and demonstrate its efficacy on several real patient outcome predictions over state-of-the-art multi-task learning schemes.

## I. Introduction

Longitudinal patient data from large scale EHR are of increasing interest for clinical hypothesis generation and testing. These datasets can contain rich information regarding the medical history of millions of patients and cover varied aspects such as diagnosis, medication, and lab tests. However, these data can be noisy, sparse, and non-uniformly distributed such that studies on a specific outcome of interest can lead to very modest number of useful patient records. These data are often insufficient to learn the underlying complex disease phenomena, especially with deep learning models. The characteristics of data imbalance often result in a 'vanishing data' problem during the analysis of specific diseases of interest with low coverage. For instance, only 210 medical records (intensive care unit stays) can be used to study patients who have experienced Rheumatic disease of heart valve and Mitral valve stenosis on the well-known MIMIC-III database [1] that originally covers 42000 records present in the complete dataset. Analysis on such datasets can be impractical and suffer from significant lack of robustness. In another case, on an EHR database spanning more than 55 million patients, querying for patients within the age group 21 – 40 and having experienced concussions that led to mild cognitive impairments returns fewer than 1100 patients. Further characterization of this cohort to focus only on patients that eventually experienced dementia takes that number below 200.

Interestingly, a significant amount of domain knowledge relating medical concepts is readily available in several medical ontologies. Phenotypes [2] corresponding to observable physical properties of patients, including diseases, have been organized in such ontologies and expressed within EHR systems. We postulate that leveraging such phenotypical relations within EHR data to jointly learn multiple outcomes in a multi-task learning setting could provide solutions to this vanishing data problem and lead to more robust models.

In this paper, we introduce **OMTL**, an ontology driven framework for multi-task learning. The main tenant of our approach is to ground the structure of the proposed multi-task framework on phenotypical medical concepts relations between patients from different cohorts. **OMTL** allows for the learning of outcomes for different phenotypes borrowing representation expressiveness from patients with similar phenotypes. This sharing allows the model to generate predictions for phenotypical cohorts even in phenotypes with small amounts of patient observations. Moreover, **OMTL** decouples the *outcome learning* from the *representation learning* of similar diseases. In this case, the model learns from patients with related diseases with different outcomes. For example, it can learn about mortality prediction for diabetic patients and readmission prediction for patients with hypertension by sharing information between diabetes and hypertension. The contributions of this paper are 3 fold:

1) We propose an approach to model different phenotypical cohorts and outcomes using domain knowledge in the form of medical concept ontological graphs.
2) We propose a novel hierarchical deep learning architecture with gating mechanisms to specialize the outcome models of each phenotype using the ontological relationship.
3) We demonstrate the efficacy of the proposed **OMTL** method on the prediction of several outcomes on patients from MIMIC-III.

In the remainder of the paper, we start in Section II with a problem description, stating the limitations of existing state-of-the-art methods and motivating the needs for structural multi-task learning followed by describing the proposed **OMTL** method in Section III. Next, in Section IV we present the experimental results geared towards the evaluation of the gain achieved by leveraging the domain ontology to relate

IBM Research, Yorktown Heights, NY USA
M. Ghalwash is also affiliated with Ain Shams University, Cairo, Egypt.

multiple tasks. Then we discuss specializing the OMTL architecture using reward shaping in Section V. We follow with an overview of recent related works in Section VI, before summarizing our conclusions in Section VII along with comments on the direction of future research. Finally, in the appendix we show additional results and more descriptions of the data used in the experiments.

## II. PRELIMINARIES AND MOTIVATION

Let us take the following example as motivation for the methodology. Assume that we want to predict in-hospital mortality (we refer to as mortality hereafter) and readmission for diabetic patients and also predict mortality for patients diagnosed with hypertension. A naive way to address this task is to construct a cohort for each individual outcome; a cohort that includes diabetic patients who did (or did not) experience mortality; another cohort consists of diabetic patients with/without readmission, and lastly a cohort of patients diagnosed with hypertension and died or survived during the hospitalization. Having limited data for each cohort reduces the opportunity to have an accurate model by learning each task individually. The multi-task framework was proposed to address issues such as these.

Recently, a multi-task mixture of experts method was proposed to share information among tasks [3]. The model has a shared representation layer, called experts, to learn different mixture of representations. These representations are combined differently for different tasks, which helps the model to share information among all tasks (outcomes) and at the same time to specialize the way these representations are aggregated for each task. However, a drawback of this approach is that it does not assume the prediction of mortality and readmission for diabetic patients are more closely related to each other (because they are related to diabetic patients) than prediction of mortality in the hypertension cohort. Moreover, different cohorts of patients (diabetes, hypertension, cardiovascular, etc.) might be have different types of relationships. For example, it is known that diabetes and hypertension are closely interlinked and are common comorbidities because of similar risk factors, whereas diabetes and atrial fibrillation share common antecedents such as hypertension. Incorporating such semantic information could lead to a better representation specialized for each phenotype, which will subsequently help the prediction of outcomes of interest for those patients who are diagnosed with that phenotype.

On top of the aforementioned issues, there may not be enough data to effectively train such a deep learning model. As was discussed in the introduction, if the user is interested in a cohort of patients developing phenotypes of narrowing of the heart valves caused by rheumatic fever: "Rheumatic disease of heart valve" and/or "Mitral valve stenosis", there are only 200 patients with these characteristics in MIMIC-III database which are not enough to learn an accurate model, even after using multi-task framework. One possible solution is to include cohorts of other phenotypes in our dataset. However, we are not particularly interested in predicting outcomes for this cohort (mortality or readmission), we are only using these data to augment our cohort of interest and learn a better predictive model.

A very recent method [4] was proposed to expand the cohort of interest by leveraging the ontology, such as SNOMED CT [5]. The ontology is represented as a directed acyclic graph over of a set of nodes (e.g. diagnosis codes or phenotypes) and edges expressing parent-child *semantic* relationships. Starting from the two *core* phenotypes the user is interested in ("Rheumatic disease of heart valve" and "Mitral valve stenosis" which are shown as black nodes in Figure 3), the method in [4] leverages the ontology to guide the selection of additional patients records from semantically-related nodes $\mathcal{G}$ (white nodes in Figure 3). Including those additional patients who have similar characteristics to the initial cohort of interest would help the model to learn better representations for the patients in the initial cohort. For example, including patients diagnosed with "Heart disease" would help the model to learn a good representation for patients diagnosed with "Rheumatic disease of heart valve" as both of these phenotypes are semantically related based on SNOMED CT ontology. After augmentation, the remaining challenge is how to include those additional patient who may not have a label, and benefit from their semantic relationship to the cohort of interest?

We propose to leverage the existing graph that relates the medical concepts to improve the phenotypical representation that can lead to more accurate predictions of outcomes specialized for each phenotype.

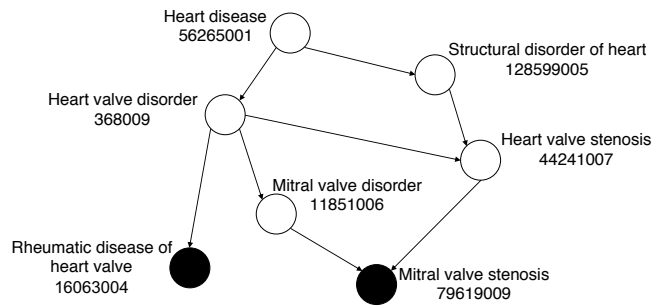

Fig. 1: A directed acyclic subgraph $\mathcal{G}$ that encompasses nodes represented in the dataset. Each node represents one medical concept. The nodes in black (i.e. core nodes) are of specific interest to the user while the white nodes are added by the augmentation process in [4]. The number in each node is the SNOMED CT code for the specified phenotype.

## III. METHODS

As discussed in the previous section, the novelty of the proposed OMTL method is in two parts. First, we decouple the representation learning of each phenotype from the prediction of the outcome of interest as to allow sharing of representations for patients with the same phenotype for different predictive tasks. Second, we impose the ontology on the network architecture in order to share information among different but related phenotypes. These two parts are described in detail in the following sections.

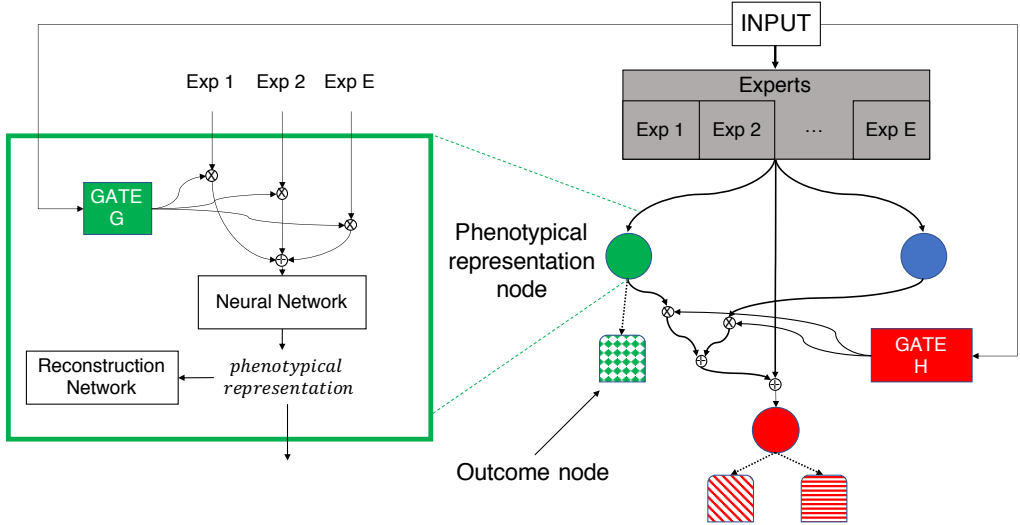

Fig. 2: OMTL: the proposed hierarchical multi-task approach. The model decouples the outcome nodes (rectangle striped nodes) from phenotypical nodes (solid colored nodes). Input of the phenotypical nodes are controlled by two gating mechanism; one to weight the inputs from phenotypical parents (Gate H) and another one to control the input from the shared representation (Gate G). Each core node is associated with different outcome nodes. The output of the outcome node is the final prediction.

Let $X = \{x_i \in \mathbb{R}^d; \forall\ i\}$ be the EHR data where $x_i$ represents data record $i$ (e.g. an admission for a patient)[1] with dimension $d$. For each record $i$, we assume the presence of all phenotypical medical concepts tied to $i$, as it is commonly done in EHR systems. For example, all diagnoses the patient developed are known at the time of admission to the intensive care unit (ICU). Using that information, we construct a subgraph $\mathcal{G}$ of the ontology with only nodes corresponding to medical concepts that are present in $X$. Any outcome prediction can be formulated as a mapping from $X$ into an outcome set $Y$. To enable sharing across outcomes, we break down this mapping as follows. We first define an encoding function $\mathcal{E}: X \to Z$, where $Z$ is the representation space. For this representation to be faithful, we enforce an unsupervised reconstruction process by learning a reconstruction function $\mathcal{R}: Z \to X$. The generated representation can be used as an input to a standard supervised machine learning algorithm $f$ to predict $Y$, $f: Z \to Y$. Learning the parameters for the mappings $\mathcal{E}$, $\mathcal{R}$ and $f$ can be achieved by minimizing $\mathcal{L}$:

$$\mathcal{L} = \underbrace{\sum_i \ell\Big(f(\mathcal{E}(x_i)), y_i\Big)}_{\mathcal{L}_1} + \lambda \underbrace{\sum_i \|\mathcal{R}(\mathcal{E}(x_i)) - x_i\|_2}_{\mathcal{L}_2} \quad (1)$$

where $\mathcal{L}_1$ is the total outcome prediction loss that aggregates the losses $\ell$ (e.g. cross entropy for classification and mean square error for regression) at the node levels corresponding to the potentially different outcomes. $\mathcal{L}_2$ is the reconstruction loss function. $\lambda \in \mathbb{R}^+$ is a hyper-parameter that weighs the importance of the two losses.

Learning a representation and a predictive model for one task is relatively straightforward by optimizing equation (1). However, learning such representations for multiple patient cohorts with different phenotypical profiles and multiple predictive tasks simultaneously is a challenging problem. We formalize OMTL by first describing a multi-task learner to decouple the outcome prediction and phenotypical representation in order to facilitate the incorporation of the hierarchical ontological structure as the second part of the model to address the aforementioned challenges.

### A. Decoupling Outcome from Phenotypical Nodes

The OMTL model is depicted in Figure 2. Input data flows from the top into a mixture of $E$ *expert nodes* responsible for clustering the input subspace to facilitate representation computations. We denote these expert nodes as $E_e(\cdot), 1 \leq e \leq E$, where each expert is a neural network producing a representation $E_e(x_i)$.

The outputs of these expert nodes are consumed by an array of $P$ phenotypical representation nodes denoted $P_p(.), 1 \leq p \leq P$. Each $P_p$ maps directly into one medical concept (i.e. the number of representation nodes is equal to the number of nodes in the subgraph $\mathcal{G}$ defined in the previous section). Each of these $P_p$ nodes ingest experts outputs via a gating function $G_p$ to learn how to combine information from all expert nodes for the generation of representations for the current medical concept. The gating network $G_p$ controls the combination of expert nodes to produce the input $\mathcal{M}_p$ to the representation node $P_p$ (described in the green box in Figure 2). These are typically simple networks that produce combination weights for the experts at each representation node:

$$G_p(x_i) = \text{Softmax}(x_i \cdot W_p + b_p) \in \mathbb{R}^E$$
$$\mathcal{M}_p(x_i) = \sum_{e=1}^{E} G_p^e(x_i) E_e(x_i) \in \mathbb{R}^{d_e} \quad (2)$$

where $W_p \in \mathbb{R}^E, b_p \in \mathbb{R}$ are parameters to be learned by the gate network during training, and $G_p^e(x_i)$ is the entry $e$ in the vector $G_p(x_i)$. The input $\mathcal{M}_p(x_i)$ is then passed to the representation node $P_p$ to produce representations

$$\mathcal{E}_p(x_i) = P_p(\mathcal{M}_p(x_i)) \quad (3)$$

for the current medical concept $p$. The objective of this part is to dynamically (using the gates) combine multiple experts for the generation of useful representation for the node $p$.

---

[1]For simplicity we assume that $x_i \in \mathbb{R}^d$ but it is straightforward to apply the proposed method to temporal data where $x_i \in \mathbb{R}^{T \times d}$

The representations $\mathcal{E}_p(.)$ are then passed as input to the *outcome nodes*. The outcome nodes leverage the representations to predict these outcomes. Each outcome node can be modeled as a neural network with the objective of predicting the outcome of interest. *It is quite important to note that OMTL can be optimized for multiple outcomes of interest as each representation node can have different and multiple outcomes*, e.g. we can associate mortality prediction for certain medical concepts and readmission prediction for another (possibly overlapping) subset of medical concepts. In Figure 2, the phenotypical red node has two associated outcome nodes. This mimics the diabetic patients in our motivating example with two prediction tasks mortality and readmission.

The model is trained in a multi-task setting allowing expert nodes and representation nodes to share learned representations and support outcome nodes in their prediction tasks. The loss function to be optimized is

$$\mathcal{L} = \underbrace{\sum_i \sum_{\substack{p \in C \\ x_i \text{ has } p}} \sum_{o \text{ assoc. with } p} \ell(f_p^o(\mathcal{E}_p(x_i)), y_i^o)}_{\mathcal{L}_1}$$
$$+ \lambda \underbrace{\sum_i \sum_{\substack{p \\ x_i \text{ has } p}} (\mathcal{R}_p(\mathcal{E}_p(x_i)) - x_i)^2}_{\mathcal{L}_2} \quad (4)$$

where $\mathcal{R}_p$ is the reconstruction network that reconstructs the input $x_i$ from the representation $\mathcal{E}_p(x_i)$, $\ell$ is the prediction loss function (e.g. cross entropy for classification) and $y_i^o$ is the label for the outcome $o$ for the visit $i$. The second sum in $\mathcal{L}_2$ runs over all representation nodes (medical concepts) that $x_i$ expresses forcing the representation from all nodes to be useful for the reconstruction of the original input. This allows the model to incorporate augmented nodes without labels. While the second sum in $\mathcal{L}_1$ runs over all *core* nodes $C$ that $x_i$ expresses, the third sum runs over all outcome nodes $o$ that are associated with the representation nodes $p$.

### B. Imposing the Hierarchical Ontology

Optimizing equation (4) can be used to solve the multi-task problem outlined above. However, depending on the choice of the phenotypes of interest, there can be a significant variation in the number of patients spanned by the representation nodes. For example, nodes corresponding to a higher level concept will typically span a much greater number of patients than the ones spanned by concepts at a leaf level of the ontology where concepts are very specialized. Interestingly, such nodes are semantically related via the medical concept ontology. Therefore, the generated representations across different medical concepts should share similar semantic relationships. OMTL aims to increase the information sharing among nodes using the ontology.

As depicted in Figure 2, the representation nodes are now connected by mirroring the ontology subgraph $\mathcal{G}$. The representation nodes without ancestors (such as the green solid node) can be computed using the same gating principle outlined in equation (2). However, for a node with ancestors (such as the red node), the model shares both the representation from the expert nodes and its parents. Assuming that $P_j$ has $n$ parents, the resultant representation $\mathcal{E}_j$ at $P_j$ is an aggregation of two vectors: (i) $\mathcal{M}_j$ computed using the combination of experts as given by equation (2) and, (ii) $\mathcal{P}_j$ computed using the combination of the representations from the parents as:

$$H_j(x_i) = \text{Softmax}(x_i \cdot W_j + b_j)$$
$$\mathcal{P}_j(x_i) = \sum_{k=1}^n H_j^k(x_i) \mathcal{E}_k(x_i) \quad (5)$$

The input to the node $P_j$ will be the aggregation of $\mathcal{M}_j$ and $\mathcal{P}_j$. The outcomes $\mathcal{E}_j(x_i)$ can be computed as:

$$\mathcal{E}_j(x_i) = P_j(\mathcal{M}_j(x_i), \mathcal{P}_j(x_i)) \quad (6)$$

Using this strategy, the ancestors of a representation node will be less affected by the 'vanishing data' problem and the structure thereby acts as a forcing mechanism to help the network to learn the representation nodes and outcome with higher efficacy. During the forward and backward pass of the training, data $x_i$ is only routed to representation nodes corresponding to phenotypes expressed in $x_i$. Algorithm 1 summarizes how OMTL process generates representations and predictions according to the proposed learning architecture.

---

**Algorithm 1:** OMTL

**Input** : data record $x_i$

1 **for** *each node $j$* **do** {nodes are sorted based on graph level}
2     **if** $x_i$ *expresses medical concept $j$* **then**
3        $\mathcal{M}_j \leftarrow$ apply equation (2)
4        $\mathcal{E}_j \leftarrow$ apply equation (3)
5        **if** *$j$ has parents* **then**
6           $\mathcal{P}_j \leftarrow$ apply equation (5)
7           $\mathcal{E}_j \leftarrow$ apply equation (6)
8        **for** *each outcome $o$ associated with node $j$* **do**
9           $\hat{y}^o = f_j^o(\mathcal{E}_j)$

---

## IV. EXPERIMENTS

In this section, we present an experimental evaluation of OMTL focusing on the impact provided by the ontology inspired architecture of OMTL that we learn within a multi-task framework. We have performed these experiments on cohorts of real patient data described in the next section. We then present details on the experimental setup before

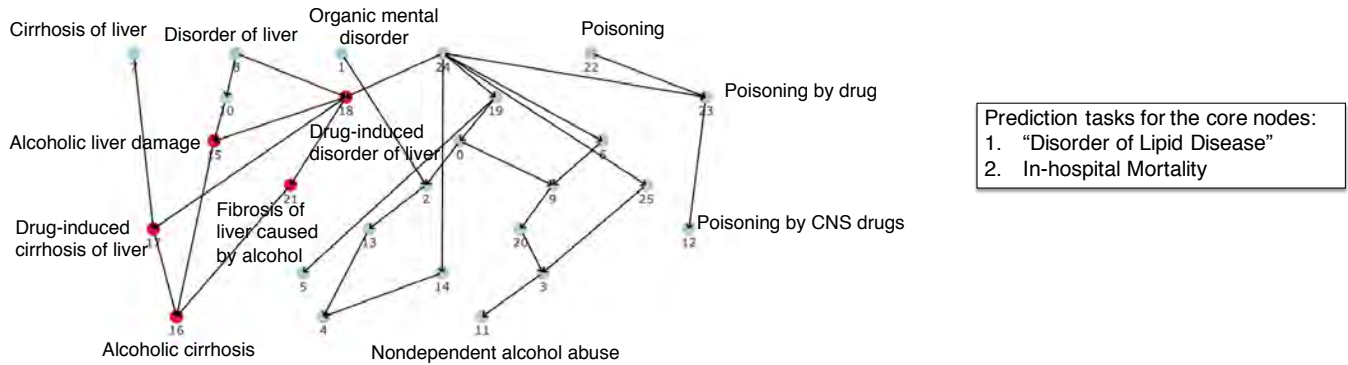

Fig. 3: Ontological graphs populated with ICU stays from MIMIC-III for the liver diseases. Core nodes are shown in red, and augmented nodes are shown in blue.

presenting the actual experimental results applying OMTL on the constructed cohorts.

### A. Data Description and Cohort Construction

We train and evaluate all models on data extracted from the publicly available MIMIC-III (Medical Information Mart for Intensive Care) database [1], a large, single-center database of patients admitted to critical care units at a large tertiary hospital. While MIMIC-III includes several facets of patient medical records, we focus our experiments on the vital signs, observations and diagnosis codes following the proposed benchmark cohort construction described in [6].

Since MIMIC-III tracks diagnosis codes in the ICD-9 format, only preserving a shallow two-level hierarchical structure of medical concepts, we map all the observed ICD-9 diagnosis codes to SNOMED CT codes, and construct a graph with a sufficiently deep hierarchy of medical concepts[2]. As a result, we obtain a complex knowledge graph where each node is associated with a set of ICU stays.

Given the knowledge graph, we have applied ODVICE [4] to obtain three ontology graphs: Liver Disorder (LD), Heart Disorder (HD), and Kideny Disorder (KD). The details of applying ODVICE are described in Appendix A. The predictive outcomes for the three graphs are phenotypes "Disorders of lipid metabolism" (DLM), "Acute myocardial infarction" (AMI), and "Chronic kidney disease" (CKD), respectively, and in-hospital mortality prediction on all cohorts. We followed [6] for processing the data and outcomes to predict. The extracted LD graph is shown in Figure 3 as an example and the other two graphs are shown in Appendix C. More statistical details on the selected cohorts can be found in Table I. The corresponding data are multivariate time series for each ICU stay. Following [6], we impute missing data and collapse time series data to single time point using the mean values of each feature and each ICU stay.

### B. Liver Graph Description

One of the prediction tasks for the core nodes in the LD graph (Figure 3) is to predict whether the patient developed

[2]https://www.nlm.nih.gov/research/umls/mapping_projects/icd9cm_to_snomedct.html

TABLE I: Statistics of datasets used for experiments. Graph name (# of nodes / # core nodes). Number of visits (number of positive labels).

| Graph | Task | # Visits (+) | Prevalence |
|---|---|---|---|
| LD (26/5) | Mortality | 4767 (837) | 0.175 |
| | DLM | 9073 (1474) | 0.162 |
| HD (21/3) | Mortality | 13067 (1867) | 0.143 |
| | AMI | 24135 (4039) | 0.167 |
| KD (78/1) | Mortality | 20316 (2749) | 0.135 |
| | CKD | 39108 (5588) | 0.143 |

"disorder of lipid metabolism" (this is not of the nodes in the given graph). The amount of data in the 5 core nodes (around 1k stays) are not enough for learning a deep learning model. When data augmentation tool ODVICE is applied to expand the 5 core node through ontology graph, the augmented cohort is expanded to have around 9k stays in total. Moreover, those additional nodes reveals related knowledge to improve the tasks on core nodes - drug/alcohol induced liver damage and disorder of liver, such as "Drug-related disorder", "Poisoning", and "Disorder of liver" that has around 5K stays. Adding additional patients who diagnosed with these phenotypes might help to learn more informed representation for patients diagnosed with the core nodes and be subsequently beneficial for predicting "disorder of lipid metabolism".

### C. Experimental Setup

As mentioned in the Methods section, our goal is to share information among outcomes at a few core nodes $C$ of interest. To realize the benefits of the augmented nodes without restricting the outcomes to be homegenous across the nodes, we train OMTL by optimizing for outcome loss at the core nodes along with reconstruction loss at all nodes. Under this setup, our primary objective is to test the hypothesis that adding the structure information from the ontology helps the model to produce more accurate predictions. To test this hypothesis, we propose to compare OMTL against multi-task baselines. Multi-task mixture of expert models [3] is the state-of-the-art multi-task model and it shares *some* of the properties of our model, such as learning a shared representation across all outcomes. Detailed difference between OMTL and [3] is described in Section VI. However, the model is optimized for supervised learning, i.e. it assumes that labels for all patients are provided, therefore the model will not benefit from the augmented data. To have a fair comparison, we modified the loss function of the baseline model by using the loss structure outlined in equation (4). This allows the model to benefit from the augmented data. This model is referred to as MMOE in the rest of this section. It uses mixture of experts [7] to share information among outcomes without using any prior knowledge about the semantic outcome-relationship. In addition, we compare the proposed method to mixture of experts (MOE) [8] where multiple experts were utilized but no gates, and the most commonly used shared-bottom (SB) multi-task deep learning structure [9] where there is only one expert.

OMTL utilizes a gating mechanism $H$ enforcing the ontological structural relationship across representation nodes. To train OMTL, we first train it without considering the hierarchy to learn the experts and their corresponding gates $G$. Then, we freeze these components and subsequently fine-tune for the outcomes and representations at the nodes by admitting the hierarchical structure.

**Implementation details**: Each module in the deep learning structure (expert, representation, reconstruction, or outcome node) is implemented as a one layer feed-forward neural network. We used 3 experts for MOE, MMOE, and OMTL. The size of the expert, reconstruction, representation modules are 41×5, 5×41, 5×5, respectively. The activation functions are Leaky ReLU for experts, Softplus for the representation layers, and ReLU for the reconstruction. Each expert has a dropout layer for regularization. The size of the gate $G_p$ is $41 \times 3$ and the size of the gate $H_j$ is $41 \times n_j$ where $n_j$ is the number of parents for the representation node $j$. We performed hyper-parameter tuning and found good performance and stability across all models by setting the batch size to 64, learning rate to 0.001, dropout to 0.5, $\lambda$ to 0.0001, and using Adam as the optimizer [10]. We evaluated the models using stratified 5-fold cross validation. the cross validation was done so as to ensure we have the same distribution of positive and negative examples in each node. All models were evaluated on exactly the same folds for a fair comparison.

*D. Multi-task Outcome Results*

We compare all baselines with OMTL using area under ROC curve (AUC-ROC). The results are shown in Table II. We also compare all models using average precision recall (APS) which summarizes the precision-recall curve; the results are shown in Appendix B. To test the significance of the OMTL model, we applied the DeLong test to test whether the ROC curves are statistically significant [11], [12]. Whenever OMTL is statistically significant (pvalue< .05) than the baseline, we superscript the baseline with *.

**Analysis:** We can see from Table II that the gain in AUC-ROC that the structure is adding to MMOE ranges from 5%-9% for the liver disorder data with ontological graph shown in Figure 3, and the gain in APS ranges from 11%-30% (Appendix B). It is worth noting that the core nodes in this graph are well-connected to each other. We hypothesize that the benefit of using the structural information among such well-connected core nodes helps OMTL to attain a significant boost in sharing information and hence provides more accurate predictions than MMOE. The ROC curves are show in Figure 4.

TABLE II: AUC under ROC for all models. * indicates when the OMTL model is statistically significant than the corresponding baseline.

| Graph | Task | SB | MOE | MMOE | OMTL |
|---|---|---|---|---|---|
| LD | Mortality | 0.715 | 0.742* | 0.715* | **0.750** |
|    | DLM | 0.572* | 0.592* | 0.576* | **0.626** |
| HD | Mortality | 0.723 | 0.700 | 0.713 | **0.729** |
|    | AMI | 0.535* | 0.537* | 0.561* | **0.593** |
| KD | Mortality | **0.728** | 0.674 | 0.697 | **0.728** |
|    | CKD | 0.743 | 0.736 | **0.749** | **0.749** |

The AUC-ROC gain in the heart disorder graph ranges from 2% to 6%, and the APS gain ranges from 6%-8%. The gain is not as significant as in the liver disorder case. A possible explanation for this is that the core nodes are less connected to each other. In such situation, the training may benefit by leveraging more outcome nodes. Similar conclusions can be reached in the kidney graph where there is only one core node and the AUC-ROC (and APS) for OMTL is similar to the MMOE model. In the next section, we discuss more this hypothesis.

Finally, since all datasets are highly imbalanced (Table I), the APS results in the appendix indicates that OMTL is able to predict the rare positive class more accurately than MMOE, which is very important in medical domain because failing to identify patients at risk is very serious since a disease has been ignored.

## V. Discussion: Specializing the OMTL Architecture

As discussed in the experimental setup, we trained the models by reducing the reconstruction loss at all nodes and optimizing for the outcome loss at only the core nodes of interest $C$. This method of training ensures that OMTL can be used in scenarios where outcomes of interest at individual representation nodes can be different. In such a training scenario, the information content at the augmented nodes is only used for learning the experts by ensuring that the model is able to reconstruct a broader variety of phenotypical information. However, when all the nodes of interest are concerned with *the same outcome*, we can fine tune the entire architecture to be discriminative w.r.t. that outcome (i.e. by reducing the outcome loss at the augmented nodes during training).

We devised an additional set of experiments to test the impact of optimizing OMTL on the core and augmented nodes together. For this set of experiments, we can control the importance of the augmented nodes towards the discriminative power of the network by weighing the losses from the nodes

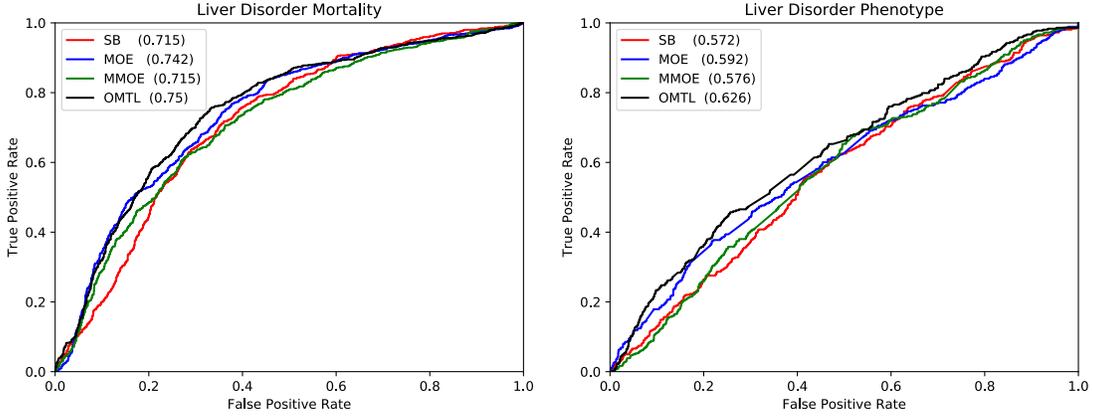

Fig. 4: ROC for liver experiments.

differently. The loss for the network for the outcome $o$ is given as:

$$\text{loss}(\mathcal{G}, o) \propto \sum_{k \in \mathcal{G}} w_k \ell(\mathbf{y_k^o}, \hat{\mathbf{y}_k^o}) \qquad (7)$$

where, $\mathbf{y_k^o}$ is the outcome $o$ for all records at the node $k$, $\hat{y}_k^o$ is the prediction, and $w_k$ is a weight to modulate the losses that control the reward the network gets for discriminating at the node $k$. There could be many ways of shaping such rewards based on a reward function $\mathcal{W}$. Intuition and domain knowledge can be used to define the right structure for $\mathcal{W}$. In this paper, we explore a linear reward scheme controlled by the level $l_k$ of the node $k$ as given below:

$$w_k = \mathcal{W}(k) \propto (\frac{l_k}{d})^f, \quad f \in (-1, 1) \qquad (8)$$

where $d$ is the depth of $\mathcal{G}$ and $f$ is a scaling factor. When $f = 1$, we reward the network for being more accurate at the leaves of $\mathcal{G}$ in contrast to rewarding the network for being accurate at the root levels when $f = -1$. $f = 0$ implies equal rewards at all nodes. After optimizing OMTL accordingly for different values of $f$, we evaluate its performance on the core nodes. We present the results of fine-tuning OMTL with reward shaping as expressed in equation (8) on Table III.

TABLE III: Specializing the OMTL model at different levels in the hierarchy. $f = 1$ rewards the network for being more accurate at the leaves. $f = -1$ rewards the network for being more accurate at the root levels. $f = 0$ implies equal rewards at all levels.

| Group | Task | $f = 1$ | $f = 0$ | $f = -1$ |
|---|---|---|---|---|
| LD | Mortality | 0.771 | **0.783** | 0.781 |
| | DLM | **0.663** | 0.652 | 0.656 |
| HD | Mortality | 0.778 | **0.782** | 0.781 |
| | AMI | 0.622 | 0.612 | **0.625** |
| KD | Mortality | 0.792 | 0.793 | **0.795** |
| | CKD | **0.766** | 0.765 | 0.765 |

As can be seen, including the outcome loss for the augmented nodes helps in specializing the OMTL model. Additionally, different forms of reward shaping can affect the performance of OMTL. For example, for the heart disorder mortality, the AUC of OMTL has been improved from 0.729 (see Table II) when optimizing the model on only the core nodes to 0.782 when optimizing the model for both core and augmented nodes. Improvements are also observed for the other experiments. This verifies our hypothesis that the connectivity of the graph assists the model in specializing for the outcome of interest.

## VI. RELATED WORK

**Multi-task learning:** Multi-task learning has received a significant amount of attention in recent years. Notably, [3] adapts a mixture-of-experts structure to jointly learn multiple outcomes by sharing a single expert, and leverages gating networks to specialize for each outcome and joint representation learning. In [13], [14], the authors employ a two-level hierarchy to regularize a learned model to share information among multiple related outcomes. [15] extended these approaches with multi-level hierarchies using a cascade approach to progressively learn outcomes at prior levels to improve learning on outcomes in subsequent levels. The model proposed in [16] learns intermediate representations for predictive tasks at multiple scales for the same input via a cascade of connected layers. In [17], the authors proposed a way to incorporate a set of *carefully* selected semantic NLP tasks into the multi-task model. "Easy" to learn tasks are supervised at the lower levels of the architecture while more complex interactions are kept at deeper layers. A soft-ordering of the shared layers was proposed in [18] to learn both layers and their order on each task.

In most cases, the shared layers are integrated in parallel order, where the outcome of one layer is fed into the next layer. OMTL is motivated by these efforts but learns phenotype specific predictive models in a multi-task setting to share information across outcomes. Most importantly, OMTL decouples the outcome nodes from the phenotypoical representation nodes. Representations learned for different phenotypes can be further specialized w.r.t. different tasks (e.g. readmission and mortality prediction) by leveraging the ontology. In addition, OMTL has the flexibility to be optimized for only a set of phenotypes in the graph (core nodes) and assign different weights to different nodes based on the domain-knowledge hierarchy.

**Knowledge graph:** Structured domain knowledge, often in graph-form, has been utilized to achieve more robust inferences over purely data-driven models. For example, [19] present a framework to enable the use of various kinds of external knowledge bases to retrieve relevant answers to a given question. It provides a graph-based model which maps text phrases to concepts in the knowledge graph using several strategies. Finally, the extracted graph is then used as an input to a neural network model for entailment.

**Outcome prediction from EHR:** Over the years, researchers have developed models on EHR data for the prediction of adverse events such as early identification of heart failure [20], readmission prediction [21], and acute kidney injury prediction [22]. Other notable works include [23] where the authors presented a multi-level attention mechanism to derive patient specific representations. The authors split the trajectory of patients into sub-sequences and use within-subsequence and between-subsequence attention levels sequentially to generate robust patient representations. Furthermore, in [6] the authors present four clinical prediction benchmarks using data derived from the MIMIC-III database, including In-hospital mortality, Decompensation, Length of stay, and Acute care phenotype classification. Additionally, they propose a deep learning model called multi-task RNN to empirically validate the four prediction benchmarking tasks. This benchmark is used as the basis for evaluating OMTL.

Recently, there has been a great interest in representing longitudinal patient data from EHR systems as low-dimensional vectors or embeddings to capture various aspects of the patient's health. One of the primary works in this field is [24] where the authors proposed a multi-layer representation learning tool for learning code and visit representations from an EHR dataset. [25] investigated the problem of data scarcity in healthcare and proposed to supplement EHR datasets with a medical ontology to provide semantic structure. The proposed attention based approach called GRAM, learns the representation of a medical concept by combining information from the concept's ancestors in the ontology. Subsequently, [26] proposed a method called MiME, to use the medical ontology to generate more robust patient embeddings by imposing a multi-level hierarchy on EHR data defined by an ontology. The levels include visits, diagnosis, and treatments.

While our proposed OMTL model also tackles the data scarcity problem by leveraging medical ontologies, it differs from these works in how the ontologies are used in learning representations. In GRAM, patient representations are learned as weighted combination of the medical concepts in the ontology. Similarly in MiME, robust patient embeddings are attained by considering the hierarchy of medical *events*. However, OMTL is aimed at learning the patient representations for multiple phenotypes in a joint multi-task manner i.e. the same patient can have different representations depending on the phenotype of interest but still benefits from the information present in more common phenotypes.

## VII. CONCLUSION

We proposed OMTL, a novel multi-task learning approach to analyze EHR data. OMTL models outcomes by incorporating knowledge from an existing ontology of phenotypical information in a multi-task setting. We have demonstrated the utility of this approach by comparing it against state-of-the-art knowledge agnostic multi-task framework over a set of experiments conducted on real patients data. Our ongoing efforts are aimed at deploying the framework for large scale EHR analysis and exploring its ability to generate reusable robust patient representations at phenotype levels. Also, the proposed method is agnostic to the EHR application, so in the future we seek to apply the method beyond EHRs in any other application where domain knowledge (or ontology) is available.

## Appendix A

Given the knowledge graph we generate three ontology graphs: Heart Disease (HD), Liver Disease (LD) and Kideny Disease (KD). For each graph, we start with a few core nodes (e.g., specific SNOMED nodes identified by domain experts for a certain phenotype) and include more SNOMED nodes to augment the ICU stays for analysis by applying the ODVICE method [4]. The data augmentation algorithm starts from this set of core nodes and grows the set by choosing a series of predecessors using an MCMC sampler with a maximum of 2 hops away from the core nodes. We apply this node growth procedure for a predefined number of iterations to obtain connected graphs with sufficient nodes and stays for training.

## Appendix B

The average precision scores (APS) results of applying OMTL and the baselines on 6 different graphs. The APS score gain of OMTL over the best baseline in 5 experiments ranges from 2.8% to 8.3%.

TABLE IV: APS using PRC for all models.

| Graph | Task | SB | MOE | MMOE | OMTL |
|---|---|---|---|---|---|
| LD | Mortality | 0.414 | 0.448 | 0.429 | **0.478** |
| | DLM | 0.086 | 0.106 | 0.084 | **0.109** |
| HD | Mortality | 0.211 | 0.221 | 0.242 | **0.262** |
| | AMI | 0.180 | 0.182 | 0.202 | **0.214** |
| KD | Mortality | 0.252 | 0.219 | 0.233 | **0.267** |
| | CKD | **0.806** | 0.798 | **0.806** | **0.806** |

## Appendix C

TABLE V: Medical concepts of liver disease graph. There are 1115 and 9081 unique stays of core nodes and entire graph respectively. Highlighted rows are the core nodes.

| No. | SNOMED | Description | # Stays |
|---|---|---|---|
| 0 | 11061003 | Psychoactive substance use disorder (disorder) | 2196 |
| 1 | 111479008 | Organic mental disorder (disorder) | 2082 |
| 2 | 11387009 | Psychoactive substance-induced organic mental ... | 982 |
| 3 | 15167005 | Alcohol abuse (disorder) | 1323 |
| 4 | 191480000 | Alcohol withdrawal syndrome (disorder) | 946 |
| 5 | 191492000 | Drug-induced delirium (disorder) | 540 |
| 6 | 191816009 | Drug dependence (disorder) | 1777 |
| 7 | 19943007 | Cirrhosis of liver (disorder) | 1988 |
| 8 | 235856003 | Disorder of liver (disorder) | 5203 |
| 9 | 2403008 | Psychoactive substance dependence (disorder) | 1323 |
| 10 | 243978002 | Liver damage (disorder) | 1115 |
| 11 | 268645007 | Nondependent alcohol abuse (disorder) | 1324 |
| 12 | 288293001 | Poisoning by CNS drugs (disorder) | 676 |
| 13 | 29212009 | Alcohol-induced organic mental disorder (disor... | 982 |
| 14 | 363101005 | Drug withdrawal (disorder) | 1088 |
| 15 | 41309000 | Alcoholic liver damage (disorder) | 1119 |
| 16 | 420054005 | Alcoholic cirrhosis (disorder) | 951 |
| 17 | 425413006 | Drug-induced cirrhosis of liver (disorder) | 951 |
| 18 | 427399008 | Drug-induced disorder of liver (disorder) | 1115 |
| 19 | 442351006 | Mental disorder due to drug (disorder) | 2502 |
| 20 | 7200002 | Alcoholism (disorder) | 1323 |
| 21 | 721710005 | Fibrosis of liver caused by alcohol (disorder) | 951 |
| 22 | 75478009 | Poisoning (disorder) | 872 |
| 23 | 7895008 | Poisoning by drug AND/OR medicinal substance (... | 832 |
| 24 | 87858002 | Drug-related disorder (disorder) | 5017 |
| 25 | 91388009 | Psychoactive substance abuse (disorder) | 1767 |

The ontology for the heart disorder and the kidney disease graph are shown in Figure 5 and Figure 6, respectively. The heart disease graph has 21 nodes, three of them are the core nodes which has 3210 unique stays. The kidney disease graph has entirely 39151 stays spanning 78 nodes, with only 1007 stays in the core node.

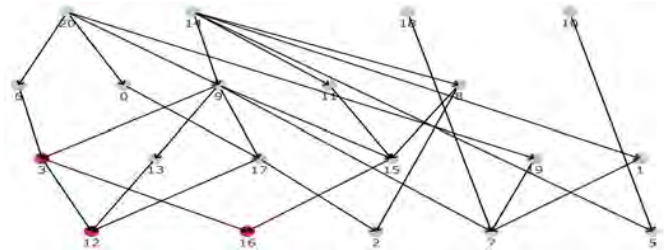

Fig. 5: Ontological graphs populated with ICU stays from MIMIC-III for the heart disorder. Core nodes are shown in red, and augmented nodes are shown in blue.

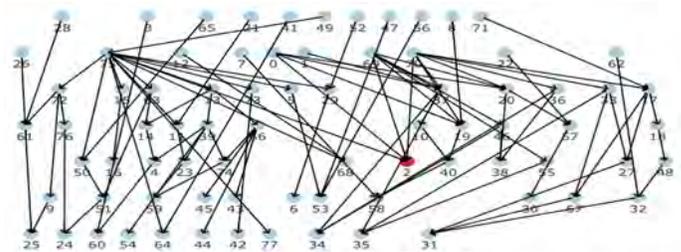

Fig. 6: Ontological graphs populated with ICU stays from MIMIC-III for the kidney disorder. Core nodes are shown in red, and augmented nodes are shown in blue.